\documentclass[runningheads]{llncs}
\usepackage{graphicx}
\usepackage{amsmath,amssymb,amsfonts}
\usepackage{algorithmic}
\usepackage{textcomp}
\usepackage{xcolor}
\usepackage{subfigure}

\usepackage{amsmath}
\begin{document}
\title{Machine Learning-based Classification of Birds through Birdsong\\}

\author{Yueying Chang\inst{1} \and
Richard O. Sinnott\inst{1}}

\institute{The University of Melbourne, Australia\\
\email{yueychang@student.unimelb.edu.au}\\
\email{rsinnott@unimelb.edu.au}}
\maketitle            
\vspace{-0.5cm}

\begin{abstract}
Audio sound recognition and classification is used for many tasks and applications including human voice recognition, music recognition and audio tagging. In this paper we apply Mel Frequency Cepstral Coefficients (MFCC) in combination with a range of  machine learning models to identify (Australian) birds from publicly available audio files of their birdsong. We present approaches used for data processing and augmentation and compare the results of various state of the art machine learning models. We achieve an overall accuracy of 91\% for the top-5 birds from the 30 selected as the case study. Applying the models to more challenging and diverse audio files comprising 152 bird species, we achieve an accuracy of 58\%.

\keywords{
Audio recognition\and data augmentation\and Convolutional Neural Networks (CNN)\and Mel Frequency Cepstral Coefficients (MFCC)}
\end{abstract}

\section{Introduction}
Animal sounds play an important role not only to help animals exchange information but also for humans to locate and understand animal activities. Researchers have shown the significance of animal sounds in areas as diverse as predicting natural disasters and markers for climate change. The advancement of modern technology makes it possible for ecologists to conveniently monitor natural environments without personally visiting  sites  using technologies for capture of large volumes of data for analysis with minimal time and cost \cite{birdCall}.

Australia is a biologically diverse country with more than a million native animal species, however, its ecological environment is especially vulnerable due to the impact of non-native species on native species as well as the direct and immediate impact of climate change. Birds are an important and effective indicator of biodiversity that often directly reflect the ecosystem health condition. Despite many advancements in Internet of Things (IoT) and artificial intelligence technologies, detecting and classifying bird species from audio recording still poses a great challenge. First, bird species are very diverse. Even within the same species, bird calls can differ depending on the season, climate and regions. Some species imitate sounds made by other species and use such sounds into their own calls \cite{sound_based}. Second, bird call recordings are typically taken in uncontrolled environments and may include background noises or  involve multiple species calling simultaneously. Third, rarer species naturally have fewer recordings available resulting in highly imbalanced data sets. Nonetheless, researchers have explored numerous methods to overcome such difficulties with machine learning based techniques showing the most promise \cite{birdCall}.

In this paper, we explore the use of machine learning approaches for audio classification of a diverse array of Australian birds. Specifically we consider the: \textit{Eastern Spinebill; Common Myna; Australian King Parrot;
Red Wattlebird; Little Wattlebird; Sulphur-crested Cockatoo; Yellow-tailed Black Cockatoo; Australian Raven; Grey Butcherbird; Laughing Kookaburra; Galah; Eastern Yellow Robin; Asian Koel; Magpie-lark; Australian Magpie; Superb Fairywren; Noisy Miner; Red-browed Finch; Crested Pigeon; Spotted Pardalote; House Sparrow; New Holland Honeyeater; Crimson Rosella; Red-whiskered Bulbul; Grey Fantail; Willie Wagtail; Spotted Dove; Pied Currawong; Rainbow Lorikeet,} and the \textit{Silvereye}.

\vspace{-0.25cm}

\section{Background and Related Work}
Many researchers have explored audio classification, covering (human) voice, music through to animal sound classification. Different pre-processing, feature extraction and classification methods have been explored. Some of the earlier works focused on music genre recognition and classification. For example, Costa et al explored music genre classification using  techniques such as extracting texture features from spectrogram images with local binary patterns (LBP), local phase quantization (LPQ) and Gabor filters, etc \cite{5977391,6623448,Costa2012MusicGC,COSTA201728}. These were fed into Support Vector Machine (SVM) models for classification. 

Image-based techniques have also been considered for animal classification tasks. Mel Frequency Cepstral Coefficients (MFCC) and Mel-frequency spectrograms were used for feature representations in acoustic modeling. Emre et al \cite{8566449} investigated animal sound classification problems by extracting MFCC features from 875 audio samples of 10 animal classes. Different optimisers were compared with Nadam giving the best overall accuracy at 75\%. Mane et al  \cite{AD} proposed a system using MFCC, Zero-Cross-Rate (ZCR) and Dynamic Time Warping (DTW) for animal audio recognition, where ZCR was used for end point detection to remove silences, and DTW  used for voice pattern classification. The combined algorithms could also be applied to recognise animal moods, states and intentions.

To resolve the huge data demands that lie at the heart of such work, several augmentation techniques were applied  to evaluate the BirdCLEF 2018 data \cite{Sprengel2016AudioBB}. This included time shift, pitch shift and adding noise. These augmentation process enlarged the data set and improved the classification performance by almost 10\%. Loris et al. \cite{app11135796} proposed six augmentation protocols including signal augmentation, spectral augmentation in different scale and time-scale modifications in different combinations.  

While some research focuses on improving classification performance, various architectures have also been considered. Shawn \cite{7952132} investigated the classification of YouTube video soundtracks by modifying some commonly used CNN pre-trained architectures to test the performance using over 5.24 million hours of video. Mangalam et al \cite{birdCall} explored the feasibility of using ResNet50 when dealing with bird call recognition by extracting spectrogram features to achieve a performance accuracy of 60\% - 72\%. Given the fact that traditional automatic animal voice detection and recognition approaches decouple segmentation and classification steps, this potentially limits the performance and propagates errors. \cite{7952135} proposed a new approach to simultaneously segment and classify bird species using an encoder-decoder architecture originally used for image segmentation.

Despite the amount of existing work on animal audio classifications, most tasks focus on distinguishing animals species in general, e.g., cats, dogs or birds, with only a limited number of works exploring individual subspecies.  Motivated by this, we aim to achieve a complete pipeline that is dedicated to address the classification problem of (Australian) birds. 

\vspace{-0.25cm}

\section{Data Preparation}
To support this work, we select a list of the top 30 urban/city dwelling birds including migratory and domestic species as defined by  \cite{australian_museum}. Raw audio data (bird calls) were retrieved from Xeno-canto  \cite{xeno-canto}. Different bird call and bird song types were used to ensure data diversity. 


\emph{LibROSA} was used for the pre-processing steps of audio files. We first extract the Mel-spectrogram for a better understanding of the hyper-parameter settings. Since most bird calls and songs have frequencies ranging between 1,000Hz-8,000 Hz, setting a fmin parameter to 1,500 effectively removes any irrelevant information that can influence the performance, while preserving the dominant features of the spectrogram. Additionally, n\_mels was set to 30 to preserve features to the greatest extent possible whilst ensuring reasonable computation resource consumption. Automatic noise reduction functions were also used. 

A feature extraction approach that directly transforms MFCC features and ZCR into numeric representations was used to create data for training. ZCR is the rate at which a signal changes sign from either positive or negative or vice versa. Raw audio is then fed into a \verb|Clip| object within specific time frames where mel-spectrogram, log amplitude and MFCC can be computed. Instead of turning MFCC features into their numeric representation, a more common way is to plot the features as images. Image representations lend themselves to comprehensive CNN architectures. We use the \emph{LibROSA} \verb|specshow| function to convert the mel-frequency spectrogram features into their associated image representation. 

Machine learning models generally perform better when data is abundant and diverse, however, manually gathering massive volumes of data is time consuming and expensive. As a result, automated data augmentation is often necessary. For an audio classification task, augmentation can be applied either to raw audio data or to the obtained images after the audio data sets are processed \cite{app11135796}. In this work we focus on augmentation of the audios files, more specifically, we apply time and frequency domain changes to the audio signals.

However the raw data  suffers from a class imbalance problem. To resolve this, three resampling strategies were applied using the \emph{imblearn} library: \textit{random downsampling} to reduce the sample number of classes; \textit{SMOTETomek oversampling} to increase the sample number of all classes to the top sampled class, and \textit{Custom strategy} to increase the minority classes and decrease the majority class sample sizes. 

The lengths of  audio files can vary considerably. \emph{LibROSA} provides functions to split audio into specific time frames. We split the audio data into 10s clips, and discard those less than 5s. 
We also introduce strides on top of time splitting to create more data. We tested 1s-3s strides within the first 100s of a given audio file. We also explore a wrap around shift approach by cutting  generated clips again by half, and placing the second part in front of the first. This approach teaches the network to deal with irregularities in the spectrogram \cite{Sprengel2016AudioBB}. It is also noted that bird calls may appear anytime in a given timeframe. \emph{LibROSA} also provides functions to split  audio signals based on non-silent intervals. An audio sample after noise reduction can be split into even smaller clips. This increases the number of samples, however, some clips with limited features are also captured, creating strangely coloured images. Such images are not beneficial to the training and were removed.

Jan \cite{Schlter2015ExploringDA} suggested that pitch shifting and time stretching can be beneficial in reducing classification errors. \emph{LibROSA} provides functions that performs pitch shifting in fractional steps. In our work, \verb|n_steps| is set to 4.

Adding background noise is another important data augmentation approach. We introduce Gaussian noise to all frequency components in an equal manner. It is noted that  Gaussian noise is easy to generate, but it may not be able to best represent real life noise conditions in some situations, hence an alternative option could be mixing audio with other real world noises.

\vspace{-0.25cm}

\section{Methodology and Results}

We first look at the approach of converting MFCC features into their numeric representation for training. Fifteen MFCCs along with the ZCR mean were extracted as features. We applied random forest, SVM and k-nearest neighbours to classify the numeric feature representations. The models were trained using five-fold cross validation. 
We also considered image classification approaches using CNNs. We compared different augmentation methods in isolation using a simple handcrafted CNN model with bird call data used as the baseline model. We  also  tested combinations of data augmentation methods. We split the raw audio files in the ratio 0.8/0.1/0.1 for training, validation and testing respectively. 

A simple handcrafted CNN was utilised. The CNN takes an input shape of 64 * 64, with two convolutional layers and two max pooling layers interspersed with each other, followed by one flattening layer and two dense layers. The model uses ReLU and sigmoid activation functions and the Adam optimiser. It was trained for 20 epochs and the accuracy of results shown in Table \ref{augmentation} below.

\begin{table}[htbp]
\footnotesize
\centering
\caption{Accuracy using a handcrafted CNN with different augmentation methods.}
\label{augmentation}
\begin{tabular}{p{4.5cm} p{1.25cm} p{1cm} p{1.8cm} p{1.8cm} p{1.8cm}}
\hline
Audio clip & Gaussian  & Filter &  No. Images & Acc. 10s clip & Acc. 5s clip\\
\hline
5s (first 100s, same below) & & & 11,265 & 0.2673 & 0.2617 \\
5s & \checkmark & \checkmark & 11,265 & 0.3352 & 0.3520 \\
5s w pitch shift & & \checkmark & 11,265 & 0.2294 & 0.2650 \\
5s origin + wrap  & & \checkmark & 11,265 & 0.2632 & 0.2762 \\
5s clip 1s stride & & & 13,095 & 0.2483 & 0.2902 \\
5s clip 2s stride & & & 15,620  & 0.2994 & 0.3310 \\
5s origin + 1s stride & & & 24,360 & 0.2697 & 0.2667\\
5s origin + 1s-2s stride & & & 39,980 & 0.2728 & 0.2854 \\
5s origin + 2s stride &  &  & 26,885  & 0.2535 & 0.3169\\
5s origin + 2s stride & \checkmark & \checkmark & 26,885 & 0.2746 & \textbf{0.3572} \\
non-silent interval & & & 32,371 & 0.1911 & 0.2004 \\
10s (whole audio, same below) & & \checkmark & 7,314 & 0.2521 & 0.1963 \\
10s origin + 2s stride & & \checkmark & 16,201 & 0.3047 & 0.2523 \\
10s &\checkmark & \checkmark & 7,314 & 0.3171 & 0.2642\\
10s origin + 2s stride & \checkmark & \checkmark & 16,169 & 0.3269 & 0.2875 \\
10s origin + 2s stride + wrap  & \checkmark & \checkmark & 23,451 & 0.3560 & 0.2798 \\
5s + 10s & \checkmark & \checkmark & 18,547 & 0.3367 & 0.3234\\
5s + 10s + 2s stride & \checkmark & \checkmark & 37,251 & \textbf{0.3759} & 0.3497\\
30s & \checkmark & \checkmark & 1,911 & 0.2452 & 0.2303\\
\hline
\end{tabular}
\end{table}

From Table \ref{augmentation} we can see almost all augmentation methods improve the accuracy. Unlike the 1D numeric data approaches, the increase in training data does not necessarily help improve the accuracy. Changes such as adding white noise or high pass filters, a similar pre-processing procedure is required for the model to produce better results. Applying vertical changes works better, while horizontal time changes add more variety to the data and they perform better when added on top of the original unprocessed data. Almost all experiments that use the Gaussian filter produce significant improvements. The non-silent interval approach performs poorly however, presumably because the clips are too short to preserve useful patterns when training, and gives rise to a large proportion of meaningless spectrograms. Considering that it takes about 60 mins to pre-process the audio files for each augmentation approach, enumerating all possible combinations is infeasible. We compute more advanced models using combinations of 5s clips with 10s clips, either with or without strides, and apply the high pass filter and white Gaussian filter to all images. The best result is achieved by combining 5s and 10s clips with a 2s stride, with both the high pass  and  white Gaussian filter. 

The CNN architecture requires the data for predictions to be in the same  format and  size as the training input to work effectively. It could be the case when an audio is split into a number of 5s clips that  predictions using these clips are in different classes. However, we are interested in the bird species within an audio file as a whole. As a result, we introduce a voting system that calculates the predictions for all  clips generated from the same audio file, and select the class with the highest vote for the final prediction. We apply data augmentation to the test data set using different combinations of  approaches to maximise the number of samples. We directly accumulate probability metrics for all classes. We found that as more variations of images were applied to the test set, the model was able to find a majority class vote for a given audio clip. 

We also explore more complicated pre-trained networks. All neural networks ran with 5s + 10s clips with a 2s stride augmentation strategy for training, and 5s clips with a 1s-3s stride augmentation strategy for validation and testing. The default input shape was 224 * 224 * 3 and the Adam optimiser and softmax activation function used. We also dynamically monitor the validation loss and conduct early stopping when the performance started to degrade. Five commonly used models of different levels of complexity trained with different epochs and applied different hyper-parameter tuning were tested: \textit{ResNet, Xception, InceptionResNet, EfficientNet} and \textit{Mobile Net}.

\begin{table}[htbp]
\footnotesize
\centering
\caption{Accuracy result for different pre-trained CNN models.}
\begin{tabular}{p{4cm} p{2cm} p{1cm} p{1cm} p{1cm} p{1cm} p{2cm}}
\hline
Model & \#params & Acc. & Prec. & Rec. & \emph{$F_1$} & Audio Acc.\\
\hline
Handcrafted CNN (baseline) & 816,958 & 0.3408 & 0.3326 & 0.3408 & 0.3196 & 0.4428 \\
ResNet50 & 26,598,302 & 0.4309 & 0.4934 & 0.4309 & 0.4229 & 0.4910\\
XCeption & 23,872,070 & \textbf{0.5448} & 0.5490 & 0.5448 & 0.5359 & \textbf{0.6627}\\
InceptionResNetV2 & 55,488,766 & 0.4946 & 0.5079 & 0.4946 & 0.4840 & 0.5994 \\
EfficientNetB3 & 13,041,478 & 0.5256 & 0.5187 & 0.5255 & 0.5008 & 0.6355 \\
MobileNetV2 & 4,139,614 & 0.4295 & 0.4373 & 0.4295 & 0.3796 & 0.4639\\
\hline
\end{tabular}
\end{table}

Xeno-canto provides raw audio in different categories. The two main ones are bird calls and bird songs. A bird call is generally simpler and shorter, and is typically produced under certain circumstances such as basic warnings and communications. Bird song tends to be longer, more musical and complex \cite{song_and_call}. In this work we primarily investigated bird calls, since we were interested in whether a model trained solely on bird calls could predict a bird from bird song. We applied the same augmentation strategies for bird songs, and applied the XCeption model and its associated hyper-parameters. As seen in Table \ref{call-song}, the model is able to distinguish bird songs better. Training and testing on bird songs alone produces an almost 10\% increase in accuracy. It is interesting to see training with bird calls and testing with bird song results in a surprisingly good accuracy of 74\%. This confirms that bird songs and bird calls are in fact a lot more similar than anticipated and can be better utilised by  mixing them together to increase the training samples thereby adding more variations to the model. 

\begin{table}[htbp]
\footnotesize
\centering
\caption{Accuracy of different categories of input training data}
\label{call-song}
\begin{tabular}{p{3cm} p{2cm} p{2cm} p{3cm}}
\hline
- & Bird Call & Bird Song & Bird Call \& Song \\
\hline
Bird Call & 0.6627 & 0.7463 & 0.7005 \\
Bird Song & 0.5301 & 0.7649 & 0.6248 \\
Bird Call \& Song & 0.7470 & 0.8731 & 0.8021 \\
\hline
\end{tabular}
\end{table}

Combining bird calls and bird songs using a fine-tuned XCeption model, achieved an accuracy of 80.21\%. However we notice that the data imbalance is still a major influencing factor. Classes with 20+ samples can obtain an accuracy over 80\% and in some cases over 90\%, whilst classes with 10- samples struggle to reach 50\% accuracy. 
The 30 bird species selected may also impact the performance, as all bird species were located in Australia with often similar habitats. Some of the species were directly related species that were hard to differentiate. The top 3 accuracy and top 5 accuracy was 86.75\% and 91.57\% respectively. Considering the limited amount of data and imbalance problem, the fine-tuned XCeption model provided a decent overall performance.

To further explore the performance of the model, we transferred the classification task to another data set - The BirdCLEF 2022 challenge \cite{birdclef}. The challenge provided 152 classes of bird recordings, with the ultimate goal of identifying 21 endangered Hawaiian bird species. 
We used similar augmentation methods as before however we discarded the 2s stride for training data. Adapting the XCeption model and training for 30 epochs, the overall audio accuracy achieved 58\%, with a top 3 accuracy 73\% and top 5 accuracy 78\%. Considering the number of classes is now five times larger than the previous Australian birds data, this result is not too disappointing, although some of the results for the endangered Hawaiian species are great, highlighting the obstacles of monitoring endangered bird populations that scientists are facing. 


\section{Conclusions and Future Work}
In this paper, we constructed a machine learning pipeline to classify a range of Australian bird species. We utilised MFCC features and Mel-frequency spectrograms extracted from the raw audio files from xeno-canto for the classification task. We achieved an ultimate audio accuracy of 80\% and a top 5 accuracy of 91\%. We identified the challenges including the limited number of real life audio data sets for endangered species that give rise to class imbalance issues which has a (negative) impact on the classification performance. We also applied our model to the BirdCLEF challenge and achieved 58\% accuracy for a 152 class classification task.

The performance of our model has some room for improvement. Apart from the problems with the data imbalance, another important thing we cannot ignore is that the collected data is recorded in real life situations on Xeno-canto. The majority of audio files are labelled with multiple secondary species whilst we only focus on the primary one. Dealing with other predictions would be beneficial for multi-label classification problems when trying to identify multiple bird species singing at the same time, although the overlap level of simultaneous sounds for different species can drastically affect the classification performance. 

The MFCC feature extraction method using SMOTE upsampling achieved an accuracy of 96\% using the random forest algorithm. Random forest outperformed others presumably because of its characteristics as an ensemble classifier. However, we believe the performance of the random forest algorithm may not always be suitable as illustrated with the BirdCLEF training data. Also, as random forest uses five-fold validation, two clips generated from the same audio with very similar patterns may fall into the training and test data sets and spuriously increase the accuracy.

\bibliographystyle{ieeetr}
\bibliography{sample}
\end{document}